# Shannon Entropy for Neutrosophic Information

**Vasile Patrascu**

Research Center for Electronics and Information Technology,

Valahia University, Targoviste, Romania.

E-mail: patrascu.v@gmail.com

**Abstract**. The paper presents an extension of Shannon entropy for neutrosophic information. This extension uses a new formula for distance between two neutrosophic triplets**.** In addition, the obtained results are particularized for bifuzzy, intuitionistic and paraconsistent fuzzy information.

## 1 Introduction

The neutrosophic representation of information was proposed by Smarandache [11], [12], [13] and it is defined by the triplet $(\mu, \omega, \nu)$ where $\mu \in [0,1]$ is the *degree of truth*, $\omega \in [0,1]$ is the *degree of indeterminacy* while $\nu \in [0,1]$ is the *degree of falsity*. The neutrosophic representation is an extension of the intuitionistic fuzzy representation proposed by Atanassov [1], [2] and also for the fuzzy representation of information proposed by Zadeh [16], [17]. We can define other parameters associated to the neutrosophic triplet $(\mu, \omega, \nu)$:

the *net truth* $\tau \in [-1,1]$, defined by:

$$\tau = \mu - \nu \tag{1.1}$$

the *bifuzzy definedness* $\delta \in [-1,1]$, defined by:

$$\delta = \mu + \nu - 1 \tag{1.2}$$

On this way, we have two systems representation of neutrosophic information: the primary space $(\mu, \omega, \nu)$ and the secondary space $(\tau, \delta, \omega)$. In addition we will define the following parameters [7], [8], [9]:

the *degree of bifuzzy incompleteness* $\pi \in [0,1]$ defined by:

$$\pi = \max(-\delta, 0) \tag{1.3}$$

the *degree of bifuzzy contradiction* $\kappa \in [0,1]$ defined by:

$$\kappa = \max(\delta, 0) \tag{1.4}$$

Between the parameters $\mu$, $\nu$, $\pi$ and $\kappa$, there exits the following relation [9]:

$$\mu + \nu + \pi - \kappa = 1 \tag{1.5}$$

There exists the following transform from the pair $(\tau, \delta)$ to the pair $(\mu, \nu)$ :

$$\mu = \frac{1 + \delta + \tau}{2} \tag{1.6}$$

$$\nu = \frac{1 + \delta - \tau}{2} \tag{1.7}$$

For the neutrosophic information $x = (\mu, \omega, \nu)$, it was defined *the complement* $\bar{x}$ by:

$$\bar{x} = (\nu, \omega, \mu) \tag{1.8}$$

After presentation of the main parameters that will be used in this approach, the next will have the following structure: section two presents a new distance for neutrosophic information; section three presents formulae for evaluating of some feature of neutrosophic information like *certainty*, *score*, *uncertainty*; section four presents the *escort fuzzy information*; section five presents the *Shannon entropy* [10] formula for neutrosophic information; section six presents the conclusion while the last is the references section.

This paper is related to my previous work [4], [5].

## 2 A distance for neutrosophic information

In this section we define a new distance for neutrosophic triplets. For two neutrosophic triplets $P_1 = (\mu_1, \omega_1, \nu_1)$ and $P_2 = (\mu_2, \omega_2, \nu_2)$, we consider the $L1$ distance $d(P_1, P_2) \in [0,3]$ define by:

$$d(P_1, P_2) = |\mu_1 - \mu_2| + |\omega_1 - \omega_2| + |\nu_1 - \nu_2| \tag{2.1}$$

The $L1$ distance [14], [15] is a metric and considering the auxiliary points $C = (1,0,1)$ and $U = (0,0,0)$ there exists the following two inequalities:

$$d(P_1, C) + d(C, P_2) \geq d(P_1, P_2) \tag{2.2}$$

And

$$d(P_1, U) + d(U, P_2) \geq d(P_1, P_2) \tag{2.3}$$

From (2.2) and (2.3) it results:

$$max\big(d(P_1, C) + d(C, P_2), d(P_1, U) + d(U, P_2)\big) \geq d(P_1, P_2) \tag{2.4}$$

We can transform (2.4) into (2.5):

$$1 \geq \frac{d(P_1, P_2)}{max\big(d(P_1, C) + d(C, P_2), d(P_1, U) + d(U, P_2)\big)} \tag{2.5}$$

The right term represents the new distance or dissimilarity, namely:

$$D(P_1, P_2) = \frac{d(P_1, P_2)}{max\big(d(P_1, C) + d(C, P_2), d(P_1, U) + d(U, P_2)\big)} \tag{2.6}$$

From (2.1) it results:

$$d(P_1, C) = 2 - \mu_1 - \nu_1 + \omega_1 \tag{2.7}$$

$$d(C, P_2) = 2 - \mu_2 - \nu_2 + \omega_2 \tag{2.8}$$

$$d(P_1, U) = \mu_1 + \omega_1 + \nu_1 \tag{2.9}$$

$$d(U, P_2) = \mu_2 + \omega_2 + \nu_2 \tag{2.10}$$

From (2.7), (2.8), (2.9), (2.10) and (1.2) it results:

$$d(P_1, C) + d(C, P_2) = 2 - \delta_1 - \delta_2 + \omega_1 + \omega_2 \tag{2.11}$$

$$d(P_1, U) + d(U, P_2) = 2 + \delta_1 + \delta_2 + \omega_1 + \omega_2 \tag{2.12}$$

From (2.11) and (2.12) it results:

$$max\big(d(P_1, C) + d(C, P_2), d(P_1, U) + d(U, P_2)\big) = 2 + |\delta_1 + \delta_2| + \omega_1 + \omega_2 \tag{2.13}$$

From (2.1), (2.13) and (2.6) it results the new distance between two neutrosophic triplets:

$$D(P_1, P_2) = \frac{|\mu_1 - \mu_2| + |\omega_1 - \omega_2| + |\nu_1 - \nu_2|}{2 + |\delta_1 + \delta_2| + \omega_1 + \omega_2} \tag{2.14}$$

From (2.1) and (2.5) it results that $D(P_1, P_2) \in [0,1]$ and we can define *the similarity* by negation:

$$S(P_1, P_2) = 1 - \frac{|\mu_1 - \mu_2| + |\omega_1 - \omega_2| + |\nu_1 - \nu_2|}{2 + |\delta_1 + \delta_2| + \omega_1 + \omega_2} \tag{2.15}$$

For $\omega = 0$, it results the particular case for *bifuzzy information* of the pair $x = (\mu, \nu)$, namely:

$$D(P_1, P_2) = \frac{|\mu_1 - \mu_2| + |\nu_1 - \nu_2|}{2 + |\delta_1 + \delta_2|} \tag{2.16}$$

$$S(P_1, P_2) = 1 - \frac{|\mu_1 - \mu_2| + |\nu_1 - \nu_2|}{2 + |\delta_1 + \delta_2|} \tag{2.17}$$

For $\omega = 0$ and $\delta \leq 0$, it results the particular case for *intuitionistic fuzzy information* of the pair $x = (\mu, \nu)$, namely:

$$D(P_1, P_2) = \frac{|\mu_1 - \mu_2| + |\nu_1 - \nu_2|}{2 + \pi_1 + \pi_2} \tag{2.18}$$

$$S(P_1, P_2) = 1 - \frac{|\mu_1 - \mu_2| + |\nu_1 - \nu_2|}{2 + \pi_1 + \pi_2} \tag{2.19}$$

For $\omega = 0$ and $\delta \geq 0$, it results the particular case for *paraconsistent fuzzy information* of the pair $x = (\mu, \nu)$, namely:

$$D(P_1, P_2) = \frac{|\mu_1 - \mu_2| + |\nu_1 - \nu_2|}{2 + \kappa_1 + \kappa_2} \tag{2.20}$$

$$S(P_1, P_2) = 1 - \frac{|\mu_1 - \mu_2| + |\nu_1 - \nu_2|}{2 + \kappa_1 + \kappa_2} \tag{2.21}$$

## 3 The certainty, the score and the uncertainty for neutrosophic information

Starting from the proposed distance defined by (2.14), we will construct some measures for the following three features of neutrosophic information: *the certainty, the score and the uncertainty*.

### 3.1 The neutrosophic certainty

For any neutrosophic triplet $x = (\mu, \omega, \nu)$ we consider its complement $\bar{x} = (\nu, \omega, \mu)$ and we define the *certainty* as dissimilarity between $x$ and $\bar{x}$, namely:

$$g(x) = D(x, \bar{x}) \tag{3.1}$$

with its equivalent form:

$$g(x) = \frac{|\mu - \nu|}{1 + |\mu + \nu - 1| + \omega} \tag{3.2}$$

In the space $(\mu, \omega, \nu)$ we identify the following properties for *neutrosophic certainty*:

(i) $g(1,0,0) = g(0,0,1) = 1$
(ii) $g(\mu, \omega, \mu) = g(\nu, \omega, \nu) = 0$
(iii) $g(\mu, \omega, \nu) = g(\nu, \omega, \mu)$
(iv) $g(\mu_1, \omega_1, \nu_1) \leq g(\mu_2, \omega_2, \nu_2)$ if $|\mu_1 - \nu_1| \leq |\mu_2 - \nu_2|$, $|\mu_1 + \nu_1 - 1| \geq |\mu_2 + \nu_2 - 1|$, and $\omega_1 \geq \omega_2$.

The property (iv) shows that the neutrosophic certainty increases with $|\tau|$, decreases with $|\delta|$ and decreases with $\omega$.

From (iv) it results that $g(\mu, \omega, \nu) \in [0,1]$ because $g(\mu, \omega, \nu) \geq g(0,1,0)$ and $g(\mu, \omega, \nu) \leq g(1,0,0)$.

In the space $(\tau, \delta, \omega)$, the *neutrosophic certainty* formula becomes:

$$g(x) = \frac{|\tau|}{1 + |\delta| + \omega} \qquad (3.3)$$

For $\omega = 0$, it results the particular case for *bifuzzy information* of the pair $x = (\mu, \nu)$, namely:

$$g(x) = \frac{|\tau|}{1 + |\delta|} \qquad (3.4)$$

For $\omega = 0$ and $\delta \leq 0$, it results the particular case for *intuitionistic fuzzy information* of the pair $x = (\mu, \nu)$, namely:

$$g(x) = \frac{|\tau|}{1 + \pi} \qquad (3.5)$$

For $\omega = 0$ and $\delta \geq 0$, it results the particular case for *paraconsistent fuzzy information* of the pair $x = (\mu, \nu)$, namely:

$$g(x) = \frac{|\tau|}{1 + \kappa} \qquad (3.6)$$

## 3.2 The neutrosophic score

From (3.2) came the idea to define the *neutrosophic score* by:

$$r(x) = \frac{\mu - \nu}{1 + |\mu + \nu - 1| + \omega} \qquad (3.7)$$

with the equivalent form in the space $(\tau, \delta, \omega)$:

$$r(x) = \frac{\tau}{1 + |\delta| + \omega} \qquad (3.8)$$

In the space $(\mu, \omega, \nu)$ the properties for the *neutrosophic score* derive from the certainty properties, namely:

(i)  $r(1,0,0) = 1; r(0,0,1) = -1$
(ii) $r(\mu, \omega, \mu) = r(\nu, \omega, \nu) = 0$
(iii) $r(\mu, \omega, \nu) = -r(\nu, \omega, \mu)$
(iv) $r(\mu_1, \omega_1, \nu_1) \leq r(\mu_2, \omega_2, \nu_2)$ if $0 \leq \mu_1 - \nu_1 \leq \mu_2 - \nu_2$, $|\mu_1 + \nu_1 - 1| \geq |\mu_2 + \nu_2 - 1|$, and $\omega_1 \geq \omega_2$
(v)  $r(\mu_1, \omega_1, \nu_1) \leq r(\mu_2, \omega_2, \nu_2)$ if $\mu_1 - \nu_1 \leq \mu_2 - \nu_2 \leq 0$, $|\mu_1 + \nu_1 - 1| \leq |\mu_2 + \nu_2 - 1|$, and $\omega_1 \leq \omega_2$

The properties (iv) and (v) show that the *neutrosophic score* increases with $\tau$. The property (iv) shows that for $\tau \geq 0$, the *neutrosophic score* decreases with $|\delta|$ and decreases with $\omega$ while the property (v) shows that the *neutrosophic score* increases with $|\delta|$ and increases with $\omega$ for $\tau \leq 0$.

From (iv) and (v), it results that $r(\mu, \omega, \nu) \in [-1,1]$ because $r(\mu, \omega, \nu) \geq r(0,0,1)$ and $r(\mu, \omega, \nu) \leq r(1,0,0)$.

For example there exists the following inequalities:

$$r(1,0,0) > r(1,1,0) > r(1,1,1) = r(1,0,1) = r(0,1,0) = r(0,0,0) > r(0,1,1) > r(0,0,1)$$

For $\omega = 0$, it results the particular case for *bifuzzy information* of the pair $x = (\mu, \nu)$, namely:

$$r(x) = \frac{\tau}{1 + |\delta|} \tag{3.9}$$

For $\omega = 0$ and $\delta \leq 0$, it results the particular case for *intuitionistic fuzzy information* of the pair $x = (\mu, \nu)$, namely:

$$r(x) = \frac{\tau}{1 + \pi} \tag{3.10}$$

For $\omega = 0$ and $\delta \geq 0$, it results the particular case for *paraconsistent fuzzy information* of the pair $x = (\mu, \nu)$, namely:

$$r(x) = \frac{\tau}{1 + \kappa} \tag{3.11}$$

## 3.3 The neutrosophic uncertainty

Finally, we define the *neutrosophic uncertainty* using the negation of certainty (3.2):

$$e(x) = 1 - \frac{|\mu - \nu|}{1 + |\mu + \nu - 1| + \omega} \tag{3.12}$$

with the equivalent form in the space $(\tau, \delta, \omega)$:

$$e(x) = 1 - \frac{|\tau|}{1 + |\delta| + \omega} \tag{3.13}$$

In the space $(\mu, \omega, \nu)$ the *neutrosophic uncertainty* verifies the following conditions:

(i)   $e(1,0,0) = e(0,0,1) = 0$
(ii)  $e(\mu, \omega, \mu) = e(\nu, \omega, \nu) = 1$
(iii) $e(\mu, \omega, \nu) = e(\nu, \omega, \mu)$
(iv)  $e(\mu_1, \omega_1, \nu_1) \leq e(\mu_2, \omega_2, \nu_2)$ if $|\mu_1 - \nu_1| \geq |\mu_2 - \nu_2|$, $|\mu_1 + \nu_1 - 1| \leq |\mu_2 + \nu_2 - 1|$, and $\omega_1 \leq \omega_2$.

The property (iv) shows that the *neutrosophic uncertainty* decreases with $|\tau|$, increases with $|\delta|$ and increases with $\omega$.

From (iv) it results that $e(\mu, \omega, \nu) \in [0,1]$ because $e(\mu, \omega, \nu) \geq e(1,0,0)$ and $e(\mu, \omega, \nu) \leq e(0,1,0)$.

For $\omega = 0$, it results the particular case for *bifuzzy information* of the pair $x = (\mu, \nu)$, namely:

$$e(x) = 1 - \frac{|\tau|}{1 + |\delta|} \tag{3.14}$$

For $\omega = 0$ and $\delta \leq 0$, it results the particular case for *intuitionistic fuzzy information* of the pair $x = (\mu, \nu)$, namely:

$$e(x) = 1 - \frac{|\tau|}{1 + \pi} \tag{3.15}$$

Formula (3.15) was obtained in [6] using other method.

For $\omega = 0$ and $\delta \geq 0$, it results the particular case for *paraconsistent fuzzy information* of the pair $x = (\mu, \nu)$, namely:

$$e(x) = 1 - \frac{|\tau|}{1+\kappa} \tag{3.16}$$

## 4 The escort fuzzy information

We will associate to any neutrosophic information $x = (\mu, \omega, \nu)$ a fuzzy one, $\hat{x} = (\hat{\mu}, \hat{\nu})$ that we will call *escort fuzzy information*. The escort fuzzy pair $(\hat{\mu}, \hat{\nu})$ will be determined in order to preserve the score of the neutrosophic triplet $(\mu, \omega, \nu)$. It will be obtained by solving the following system:

$$\hat{\mu} + \hat{\nu} = 1 \tag{4.1}$$

$$\hat{\mu} - \hat{\nu} = r(\mu, \omega, \nu) \tag{4.2}$$

It results the following values for the escort fuzzy pair $(\hat{\mu}, \hat{\nu})$:

$$\hat{\mu} = \frac{1+r}{2} \tag{4.3}$$

$$\hat{\nu} = \frac{1-r}{2} \tag{4.4}$$

with the following equivalent forms:

$$\hat{\mu} = \frac{\mu + \pi + \frac{\omega}{2}}{1 + |\delta| + \omega} \tag{4.5}$$

$$\hat{\nu} = \frac{\nu + \pi + \frac{\omega}{2}}{1 + |\delta| + \omega} \tag{4.6}$$

For $\omega = 0$, it results the particular case for *bifuzzy information* of the pair $x = (\mu, \nu)$, namely:

$$\hat{\mu} = \frac{\mu + \pi}{1 + |\delta|} \tag{4.7}$$

$$\hat{\nu} = \frac{\nu + \pi}{1 + |\delta|} \tag{4.8}$$

For $\omega = 0$ and $\delta \leq 0$, it results the particular case for *intuitionistic fuzzy information* of the pair $x = (\mu, \nu)$, namely:

$$\hat{\mu} = \frac{\mu + \pi}{1 + \pi} \tag{4.9}$$

$$\hat{\nu} = \frac{\nu + \pi}{1 + \pi} \tag{4.10}$$

For $\omega = 0$ and $\delta \geq 0$, it results the particular case for *paraconsistent fuzzy information* of the pair $x = (\mu, \nu)$, namely:

$$\hat{\mu} = \frac{\mu}{1+\kappa} \tag{4.11}$$

$$\hat{\nu} = \frac{\nu}{1+\kappa} \tag{4.12}$$

The escort fuzzy pair $(\hat{\mu}, \hat{\nu})$ can be used to extend existing results from fuzzy theory [16], [17] to its extensions. In this paper we will use the escort fuzzy pair for extending the Shannon fuzzy entropy to Shannon neutrosophic entropy.

## 5 The Shannon entropy for neutrosophic information

For any neutrosophic triplet $x = (\mu, \omega, \nu)$, using the escort fuzzy pair $\hat{x} = (\hat{\mu}, \hat{\nu})$ we will define the Shannon entropy $E_S$ for neutrosophic information using the following formula:

$$E_S(x) = e_S(\hat{x}) \tag{5.1}$$

where $e_S$ represents the Shannon fuzzy entropy. For a fuzzy information $\mu$, De Luca and Termini [3] extended the Shannon formula for calculating the fuzzy entropy by:

$$e_S(\mu) = -\mu \ln(\mu) - (1-\mu)\ln(1-\mu) \tag{5.2}$$

From (5.2) it results:

$$e_S(\hat{x}) = -\hat{\mu}\ln(\hat{\mu}) - \hat{\nu}\ln(\hat{\nu}) \tag{5.3}$$

From (4.5), (4.6), (5.1) and (5.3) it results the Shannon variant for neutrosophic entropy:

$$E_S(x) = -\frac{\mu + \pi + \frac{\omega}{2}}{1 + |\delta| + \omega} \ln\left(\frac{\mu + \pi + \frac{\omega}{2}}{1 + |\delta| + \omega}\right) - \frac{\nu + \pi + \frac{\omega}{2}}{1 + |\delta| + \omega} \ln\left(\frac{\nu + \pi + \frac{\omega}{2}}{1 + |\delta| + \omega}\right) \tag{5.4}$$

There are the next three equivalent formulae:

Using (4.3) and (4.4) it results:

$$E_S(x) = -\frac{1+r}{2}\ln\left(\frac{1+r}{2}\right) - \frac{1-r}{2}\ln\left(\frac{1-r}{2}\right) \tag{5.5}$$

Because in (5.5) there exists symmetry between $r$ and $-r$, it results:

$$E_S(x) = -\frac{1+|r|}{2}\ln\left(\frac{1+|r|}{2}\right) - \frac{1-|r|}{2}\ln\left(\frac{1-|r|}{2}\right) \tag{5.6}$$

From (3.2), (3.3), (3.7) and (5.6) it results:

$$E_S(x) = -\frac{1+g}{2}\ln\left(\frac{1+g}{2}\right) - \frac{1-g}{2}\ln\left(\frac{1-g}{2}\right) \tag{5.7}$$

We notice that:

$$\frac{\partial g}{\partial |\tau|} = \frac{1}{1 + |\delta| + \omega} \tag{5.8}$$

$$\frac{\partial g}{\partial |\delta|} = -\frac{|\tau|}{(1 + |\delta| + \omega)^2} \tag{5.9}$$

$$\frac{\partial g}{\partial \omega} = -\frac{|\tau|}{(1 + |\delta| + \omega)^2} \tag{5.10}$$

$$\frac{\partial E_S}{\partial g} = \frac{1}{2}\ln\left(\frac{1-g}{1+g}\right) \tag{5.11}$$

$$\frac{\partial E_S}{\partial |\tau|} = \frac{1}{2}\frac{1}{1 + |\delta| + \omega}\ln\left(\frac{1-g}{1+g}\right) \tag{5.12}$$

$$\frac{\partial E_S}{\partial |\delta|} = -\frac{1}{2}\frac{|\tau|}{(1 + |\delta| + \omega)^2}\ln\left(\frac{1-g}{1+g}\right) \tag{5.13}$$

$$\frac{\partial E_S}{\partial \omega} = -\frac{1}{2}\frac{|\tau|}{(1 + |\delta| + \omega)^2}\ln\left(\frac{1-g}{1+g}\right) \tag{5.14}$$

because

$$\frac{1-g}{1+g} \leq 1$$

it results:

$$ln\left(\frac{1-g}{1+g}\right) \leq 0 \tag{5.15}$$

From (5.12), (5.13), (5.14) and (5.15) it results:

$$\frac{\partial E_S}{\partial |\tau|} \leq 0 \tag{5.16}$$

$$\frac{\partial E_S}{\partial |\delta|} \geq 0 \tag{5.17}$$

$$\frac{\partial E_S}{\partial \omega} \geq 0 \tag{5.18}$$

As conclusion, it results that the Shannon entropy for neutrosophic information defined by (5.4) verifies the condition (iv) from section 3.3, namely it decreases with $|\tau|$, increases with $|\delta|$ and increases with $\omega$.

Also the function $E_S$ defined by (5.4) verifies the conditions (i) and (iii). In order to verify the condition (ii) it necessary to multiply by the well-known normalization factor:

$$\lambda = \frac{1}{\ln(2)} \tag{5.19}$$

Finally it results the normalized variant for Shannon entropy, namely:

$$E_{SN}(x) = -\frac{1}{\ln(2)}\left(\frac{\mu + \pi + \frac{\omega}{2}}{1 + |\delta| + \omega} ln\left(\frac{\mu + \pi + \frac{\omega}{2}}{1 + |\delta| + \omega}\right) + \frac{\nu + \pi + \frac{\omega}{2}}{1 + |\delta| + \omega} ln\left(\frac{\nu + \pi + \frac{\omega}{2}}{1 + |\delta| + \omega}\right)\right) \tag{5.20}$$

For $\omega = 0$, it results the Shannon entropy formula for *bifuzzy information* of the pair $x = (\mu, \nu)$, namely:

$$E_{SN}(x) = -\frac{1}{\ln(2)}\left(\frac{\mu + \pi}{1 + |\delta|} ln\left(\frac{\mu + \pi}{1 + |\delta|}\right) + \frac{\nu + \pi}{1 + |\delta|} ln\left(\frac{\nu + \pi}{1 + |\delta|}\right)\right) \tag{5.21}$$

For $\omega = 0$ and $\delta \leq 0$, it results the Shannon entropy formula for *intuitionistic fuzzy information* of the pair $x = (\mu, \nu)$, namely:

$$E_{SN}(x) = -\frac{1}{\ln(2)}\left(\frac{\mu + \pi}{1 + \pi} ln\left(\frac{\mu + \pi}{1 + \pi}\right) + \frac{\nu + \pi}{1 + \pi} ln\left(\frac{\nu + \pi}{1 + \pi}\right)\right) \tag{5.22}$$

For $\omega = 0$ and $\delta \geq 0$, it results the Shannon entropy formula for *paraconsistent fuzzy information* of the pair $x = (\mu, \nu)$, namely:

$$E_{SN}(x) = -\frac{1}{\ln(2)}\left(\frac{\mu}{1 + \kappa} ln\left(\frac{\mu}{1 + \kappa}\right) + \frac{\nu}{1 + \kappa} ln\left(\frac{\nu}{1 + \kappa}\right)\right) \tag{5.23}$$

## 6 Conclusions

In this paper, we presented a new formula for calculating the distance and similarity of neutrosophic information. Then, we constructed measures for information features like *score, certainty and uncertainty*. Also, a new concept was introduced, namely *fuzzy information escort*. Then, using the fuzzy information escort, Shannon's formula for neutrosophic information was extended. It should be underlined that Shannon's entropy for neutrosophic information verifies the four defining conditions of neutrosophic uncertainty. Also, all the obtained results for the triplet $(\mu, \omega, \nu)$ were particularized for bifuzzy information, intuitionistic fuzzy information and paraconsistent fuzzy information of the pair $(\mu, \nu)$.